# Feature Learning with Multi-Stage Vision Transformers on Inter-Modality HER2 Status Scoring and Tumor Classification on Whole Slides

Olaide N. Oyelade[1, *], Oliver Hoxey[2], and Yulia Humrye[2]

onoyelade@ncat.edu, 2410845@stu.chi.ac.uk, y.humrye@chi.ac.uk

[1]Department of Computer Systems Technology, North Carolina A&T State University, United States of America
[2]School of Nursing and Allied Health, Biomedical Sciences, University of Chichester, United Kingdom
[3]Yale School of Medicine, Yale University

*Corresponding Author: **onoyelade@ncat.edu**

**Abstract**
The popular use of histopathology images, such as hematoxylin and eosin (H&E), has proven to be useful in detecting tumors. However, moving such cancer cases forward for treatment requires accurate on the amount of the human epidermal growth factor receptor 2 (HER2) protein expression. Predicting both the lower and higher levels of HER2 can be challenging. Moreover, jointly analyzing H&E and immunohistochemistry (IHC) stained images for HER2 scoring is difficult. Although several deep learning methods have been investigated to address the challenge of HER2 scoring, they suffer from providing a pixel-level localization of HER2 status. In this study, we propose a single end-to-end pipeline using a system of vision transformers with HER2 status scoring on whole slide images of WSIs. The method includes patch-wise processing of H&E WSIs for tumor localization. A novel mapping function is proposed to correspondingly identify correlated IHC WSIs regions with malignant regions on H&E. A clinically inspired HER2 scoring mechanism is embedded in the pipeline and allows for automatic pixel-level annotation of 4-way HER2 scoring (0, 1+, 2+, and 3+). Also, the proposed method accurately returns HER2-negative and HER2-positive. Privately curated datasets were collaboratively extracted from 13 different cases of WSIs of H&E and IHC. A thorough experiment is conducted on the proposed method. Results obtained showed a good classification accuracy during tumor localization. Also, a classification accuracy of 0.94 and a specificity of 0.933 were returned for the prediction of HER2 status, scoring in the 4-way methods. The applicability of the proposed pipeline was investigated using WSIs patches as comparable to human pathologists. Findings from the study showed the usability of jointly evaluated H&E and IHC images on end-to-end ViTs-based models for HER2 scoring.

**Keywords:** Vision transformer, convolutional neural network, hematoxylin and eosin (H&E), human epidermal growth factor receptor 2 (HER2), breast cancer,

## 1. Introduction

Cancer detection and grading are being achieved using hematoxylin and eosin (H&E) images, in addition to other medical image modalities. The H&E images are clinical standard with stain, which helps to reveal the hidden cell structure. The dye stain provides patterns that pathologists can observe crucial morphological information through the cell structure, thereby uncovering necessary details for cancer diagnosis. On the other hand, human epidermal growth factor receptor 2 (HER2) is a protein expressed that acts as an important enabling further understanding of detected breast cancer. While HER2 is not the only biomarker supporting this process, other biomarkers such as Ki-67, PgR, and ER are also useful in informing the treatment pathway and application of HER2-targeted therapy for diagnosed breast cancer cases [1]. Although H&E exposes the morphological structure in the histopathology tissue, its stain does not directly supply the HER2 status, which guides the administration of treatment; it provides a comparative option with immunohistochemistry (IHC) stained images. Hence, both H&E and IHC-stained whole slide images (WSIs) are comparatively observed for assessment of the amount of HER2 protein expression to inform HER2 status [2]. Hence, IHC and HER2 gene amplification within situ hybridization (ISH) or fluorescence in situ hybridization (FISH) are crucial for breast cancer subtyping. Since IHC exploits the bond between antigen and antibody to expose abnormal antigen in tissue, what remains challenging is corresponding this with the morphological patterns seen in H&E. This challenge is often seen in intra- and inter-observability issues [3] [4] [5] when a pathologist compares an H&E slide with an IHC-stained slide for IHC-scoring. This, therefore, implies that the score, often categorized as 0 or 1+ (HER2-negative), 2+ (equivocal/borderline), and 3+ (HER2-positive), can be misleading due to the observability challenge. In addition to the scoring accuracy problem, the difficulty of correctly classifying the scores in the binary form HER2-negative/ HER2-positive (0, 1+, and 2+ / 2+,3+), 3-way form (0, low and high), and 4-way, leads to ineffectiveness in levels of HER2 expression [6].

In view of the human-introduced inaccuracy problems in staging of HER2 expression, and the huge cost of annotating WSIs, deep learning algorithms have evolved to improve the process. Moreover, the emergence of the relevance of digital image analysis in enabling the scoring or status of HER2 [7], has widened opportunities to automate the process. For instance, deep learning models have been trained to intelligently predict HER2 status from H&E [1] as against the pathologist-led comparative method of H&E and IHC WSIs. Another interesting perspective on this is the application of generative models to virtually generate IHC-stained H&E images [8] to improve the effectiveness of deep learning models for HER2 status prediction. Prominent and promising among these deep learning methods are the vision-language models [9], also known as vision transformers (ViTs). As a matter of fact, the ViTs architectures have been widely used for tumor classification of H&E WSIs [10] [11] [12], and more advanced methods using ViTs [13] [14] [15] for classification or grading malignant tissues. Meanwhile, traditional deep learning architectures like the convolutional neural methods (CNN) have also demonstrated good performance [16] [17], and their hybrid with ViTs [18]. Unfortunately, addressing heterogeneity pattern representation of tumors on WSIs, very high resolutions of WSIs, and accurately predicting HER2-low expression (0 and 1+) using ViTs remains challenging [19]. Recent advances on ViTs, such as hierarchical transformer with shifted window [20] [21], hierarchical self-attention [22], using visual pretext task on ViTs [23], have been proposed for improving computational pathology. Moreover, these advances notwithstanding, there is a growing concern about how these recent solutions can accurately predict molecular information merely from histomorphology.

Several attempts have been made to adapt deep learning architectures to HER2 status prediction, either using H&E samples or directly from IHC-stained WSIs. For instance, CNN architectures have been jointly applied to H&E in HER2 [24], while the difficulty of heterogeneous staining in the IHC-stained WSIs has also been investigated [25]. Furthermore, the challenging task of isolation of features supporting HER2 status prediction has been reported in [5] [26]. In addition, hybrids of neural networks with other techniques such as rule-based, multi-instance learning (MIL), and probabilistic [27] [28] [29] [30] have been adapted for HER2 binary, 3-way, and 4-way classification. As mentioned in the previous paragraphs, addressing the limitation of ViTs related to pathological computation is promising. Hence, correlational attention with ViTs [31], contrast-enhanced MRI (CE-MRI) with ViTs [32], and Fourier transform with ViTs [33]. However, central to the challenge of predicting the score of HER2 is high resolution scoring system [34], improving the performance to accurately classify the low-expressing status of HER2 breast cancer in molecular biomarkers such as IHC and FISH. Also, pathologists' inherited biases in scoring HER2 0, 1+, and 2+/FISH remain a critical issue impeding discrimination against which breast cancer cases proceed to receive HER2-therapy [35]. In addition to these, there are challenges of inconsistency and heterogeneous morphology patterns in IHC-stained WSIs since IHC evaluates HER2 staining [36]. Uncertainty in the usability of weakly-supervised and unsupervised feature learning methods remains a challenge to scoring low (0 and 1+) HER2 status in both H&E and IHC-stained WSIs. Although recent research has shown significant progress in addressing some of these problems, we discovered that there are still substantial issues left unaddressed among these problems. To close these research gaps, the methods described in this study are carefully investigated and reported.

In this study, we leveraged a system vision transformer to ensure H&E and IHC WSIs are comparatively screened for malignancy and scoring of HER2 protein expression. This pipeline first addresses the notorious problem of computational demand of processing WSIs for both classification and HER2 status prediction. Moreover, rather than approaching the score-prediction from the angle of classification, we formulated the problem of HER2 scoring as a reflection of the clinical procedures. We trained a ViT-based segmentation model to accurately identify and localize different degrees of staining, indicating the 4-way subtyping of HER2 status. Furthermore, a novel mapping function is introduced to ensure H&E WSIs are correlated with a corresponding IHC-stained WSIs, which naturally do not have any correspondence. Lastly, returning a single score for a WSIs can be challenging, and localizing tumor location across the high-resolution WSIs is hectic. This study addresses these two by automating the transitions of WSIs-to-patch-WSIs through a novel mechanism. The proposed solution in the study adequately addresses the recurring problem of accurately predicting HER2 scores at the lower levels (0, 1+, and 2+) [37] and at the HER2 positive/higher levels, and the former amounting to over 50% of breast cancer cases [38]. The following summarizes the technical contribution of this paper:

1. **H&E Whole Slide Image Classification:** We implemented a system to automatically apply WSIs to detect localized regions with abnormality, leading to cancer detection.

2. **Mapping Regions in H&E into Regions in IHC WSIs:** A novel approach is proposed to correspond the tumor-defected regions of H&E tissue with an exact region in the matching IHC WSIs.

3. **Semantic Pixel Segmentation and Annotation:** We adapted a segmentation transformer to accurately achieve pixel-level categorization and segmentation of different classes of stains, suggesting HER2 status

4. **Pipeline for H&E classification and HER2 Scoring:** Two separate ViT-based models were proposed for the classification task on tumor identification and HER2 stain-intensity-status computation.

5. **Experimentation on Privately Curated Dataset:** We curated datasets for thorough experimentation on the complete pipeline and carried out comparative analysis with the current state-of-the-art (SOTA).

The remainder of this paper is structured as follows: Section 2 discusses related works, Section 3 presents the proposed methodology, Section 4 outlines the experimental setup and datasets, Section 5 presents results and discussion, and Section 6 concludes the study with key findings and future directions.

**2. Related Works**
To provide a clear context for the approach proposed in this study, a detailed review of related works is presented in this section. First, some very similar methods based on CNN and other machine learning algorithms were considered. Furthermore, furthermore, a detailed review on other studies utilizing ViT-based models were explored and summarized.

Several deep learning network architectures have been reported to yield interesting performance for HER2 scoring. The work of [24] is a good demonstration of the applicability of deep learning model for automated scoring of HER2 despite the heterogeneous staining in the images. Likewise, CNN has also been reported in [25] to return good performance on manually extracted patches of H&E in HER2 status detection. Similarly, [5] also deployed a pyramid sampling technique which is based on deep learning to ensure spatial features are adequately extracted for understanding the status HER2 expression. This motivation is related to what is described in [26] for feature extraction, except that a re-parameterization deep learning approach was used. While those two were based on automating the HER2 scoring, the report in [39] suggested a semi-automated process that relies on a two-stage deep learning model for both segmentation of patches and classification. The two-stage approach provides an interesting solution to handling WSIs for scoring HER2 expression and in estimating the percentage of degradation in the slide. Rather than simply classify HER2 expression into the four major groups, [28] the combined Xception model and Grad-CAM are used to ensure that explainability functionality is added to the pipeline. As a result, HER2 scoring was achieved through classification into HER2-0, HER2-1+, HER2-2+, and HER2-3+ using H&E images. Contrary to the popular use of supervised learning, authors in [40] proposed adapting weakly supervised learning by leveraging momentum contrastive learning (MoCo-v2) for feature extraction from H&E WSIs. However, the argument by [41] on the effectiveness of supervised method using neural network when applied directly to H&E images, shows that the unsupervised method is competitive. On the contrary, weakly supervised multi-modal contrastive learning has been reported for ensuring that feature information from separate modalities is effectively converged for HER2 scoring [27]. The multimodality issue ensures that features embedded in H&E and IHC WSIs are collectively interpreted through an attention mechanism with a ranked HER2 score. Meanwhile, a system of deep learning models has been integrated to simplify the classification of HER2 status into negative, low, and high expression [42]. However, most studies have largely focused on the benchmarked 4-way classification, which includes the 0, 1+, 2+, and 3+; hence, our focus shall be on this 4-way approach. On the other hand, techniques such as transfer learning, multi-instance learning (MIL), rule-based, and probabilistic aggregation (TL-PA) have been reported by [29] [30] to work in HER2 scoring when applied to H&E WSIs. However, the results obtained by these methods are not as competitive as the deep learning models highlighted by previous studies. This demonstrates that the deep learning methods are foundational and can motivate new architectures.

Considering the current progress using deep learning models, we explored the performance of vision transformers and emphasized the methodology applied in related studies. For instance, [43] have investigated the performance comparison of traditional neural networks with vision transformers in computing HER2 status. The experimentation on WSIs was focused on measuring the performance of DenseNet201, GoogleNet, and

MobileNet, compared with ViT model. Results obtained showed the viability of the transformer-based technique. The work in [31] has also promoted the approach of combining ViTs with a correlational attention neural network (Corr-A-Net), which is supported by an estimation neural network for confidence score computation. Similarly, [32] have also reported the benefit of integrating dynamic contrast-enhanced MRI (DCE-MRI) with ViTs for improved HER2 expression estimation. A demonstration of the usefulness of a transformer-based approach for HER2 scoring was reported in [44]. Using H&E WSIs, feature learning was carried out using a kernel attention transformer (KAT) to ensure hierarchical context information is derived. The study applied the feature learning model on patches obtained from WSIs using Pathology Language and Image Pre-Training (PLIP). Another interesting application of ViT to the task of HER2 scoring has been reported in [33] where the Fourier transform was interfaced with a transformer model for the classification of HER2 status. The use of spatial transformers for the localization of features suggesting abnormality for detecting HER expression computation has been investigated in [45]. Authors noted that once the features are localized, all hotspots in WSIs are spatially aligned using another ViT model, while relying on a loss computation to HER2 status. ViT models have been investigated on molecular data often located in IHC assays (Ki67, ER, PR, and HER2) for status prediction. Hence, contrary to the focus on HER2 status classification, the work in [46] adapted ViTs for Ki67, ER, PR, and HER2 status on a 3-way classification. Similarly, authors in [47] have demonstrated that it was much better to focus on the use of ViT models for evaluating the potential benefit of neoadjuvant therapy on already positive HER2 status, a deep learning model, and apply the model to predict the efficacy of neoadjuvant therapy. The investigation was supported by multimodal data involving IHC WSIs and next-generation sequencing (NGS). This demonstrates the possibility of widening the use of ViT-based models for both classification and evaluation of the effectiveness of the application of some recent cancer drugs in the treatment of new cases.

To summarize the review, it is worth noting that the ViT-based approach outperforms the traditional deep learning architectures. In this study, we intend to address the limitations of ViT-based methods to ensure that a complete pipeline for the computation of the HER2 status is accurate, applicable, and reflects the actual clinical outcome based on professional guidelines. This is also aimed at identifying candidate WSIs and cases relevant to be tested on novel antibody-drug conjugates.

## 3. Methodology

The approach proposed in this study is detailed in this section. First, the end-to-end pipeline for H&E and IHC WSIs classification and scoring is provided in detail. This pipeline gives an overview of the proposed method and underscores the motivation for the study. Secondly, the mapping mechanism for corresponding localizing regions from H&E into IHC-stained images is also presented. We followed this with a description of how HER2 scores are assigned from tile-level to WSI-level. Moreover, the system of transformer-like architectures applied for tumor and stain classification is discussed while providing clarification on how the score-stain regions are segmented in the WSIs.

### 3.1 End-to-end transformer-based pipeline for HER2 status scoring

The novelty of this study is centered on the proposed end-to-end pipeline discussed in this section, as seen in Figure 1. The figure shows an illustration of how subsystems interface together and were trained as an end-to-end solution for HER2 status scoring. The pipeline shows four major compartments, namely: image annotation, the dataset creation and preprocessing, a system of transformer-like architectures, and the detection and scoring components. The image annotation phase was human-driven since pathologists were mainly relied upon to carefully prepare the WSIs for all cases applied in this study. Specifically, the pathologist obtained the slides and stained H&E and IHC samples for the purpose of this research. Once this was completed, the WSIs were passed as input into the end-to-end pipeline. Tiles/patches were then generated for a desired window size $WxH$ (e.g, 512x512). In the dataset creation and preprocessing phase, curated tiles were correctly assigned labels to enable supervised training using two separate transformer architectures (section 3.4). Tiles from both H&E and IHC-stained WSIs were automatically extracted, and their corresponding labels were assigned. On the other hand, we leveraged Segment.AI to create a dataset for a transformer-based segmentation model (section 3.5). Furthermore, all models, transformer architectures described in sections 3.4 and 3.5, were trained for feature extraction and for other downstream tasks – classification and segmentation. In the detection and scoring, samples of H&E and IHC tiles are passed through the trained model for an end-to-end process involving classification, scoring (see section 3.3), and correspondingly mapping of tiles from H&E into IHC (see section 3.2) to accurately project a view of malignant regions with patch-based annotated scores.

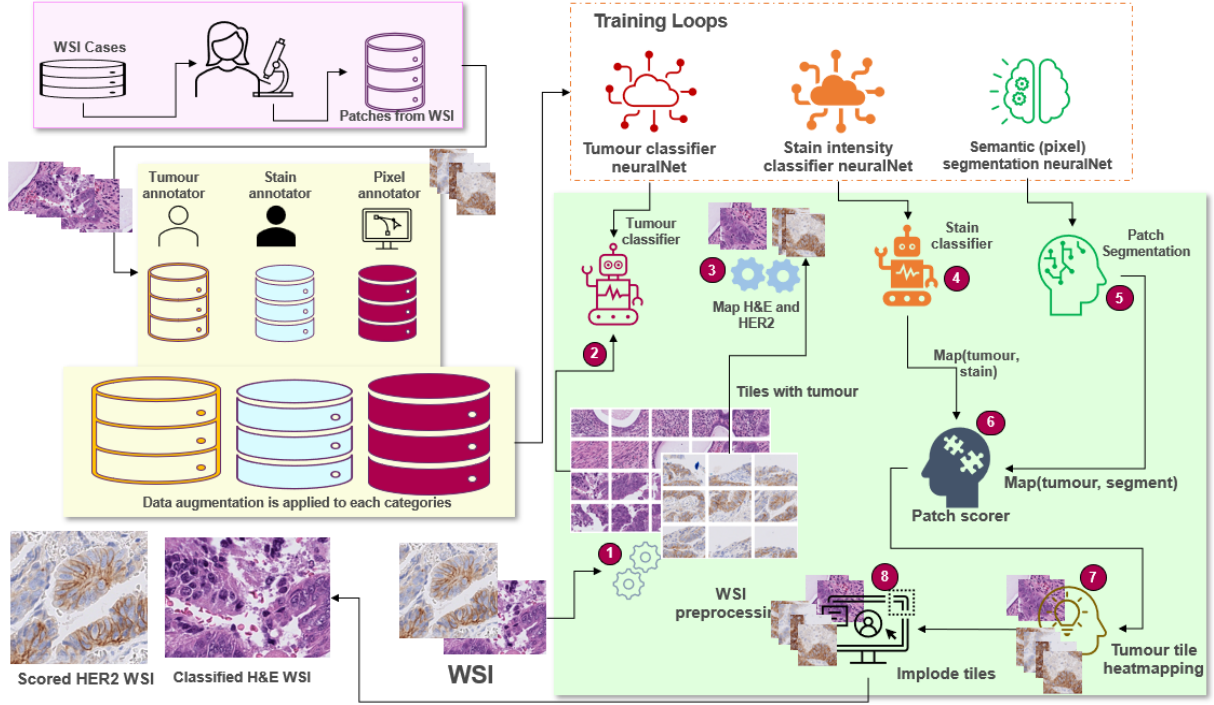

**Figure 1**: An end-to-end pipeline utilizing multiple/system of vision transformer for H&E classification status scoring of HER2 using whole slide images of digital histopathology

Output from the pipeline is an imploded tiles transformed into well processed WSIs of H&E and IHC. The aim is to provide pathologists with an assessment and verification tool enabling HER2 status prediction. In the following subsections, component-wise description of the pipeline is discussed.

### 3.2 Inter-modality image mapping

H&E and IHC WSIs may not often clearly indicate any correlation in their respective stains. However, it is necessary to map regions in these different modalities to explain the detection of malignancy and the HER2 scoring. As a result, the inter-modality mapping function is described. It relies on $g$ function which convolve an $f x f$ filter (e.g 512x512) on the WSIs of the two separate modalities. This is illustrated by equation (1) where $P_R(S)$ accepts the whole slide $S$, and $g(.)$ which is like a $\odot$ operator finding all the possible tiles of size $f x f$ across rows and columns of $S$.

$$P_R(S) = \sum_{i=o}^{R} g(S, C)_i \quad (1)$$

Note that $C = \lceil W/f \rceil$ and $R = \lceil H/f \rceil$. However, each $P_R(S)$ is more of a block in $S$, is therefore further explained in equation (2).

$$P_R(S) = \{p_{ij} \mid i = 0 \dots R, and\ j = 0 \dots C\} \quad (2)$$

Hence, the same operation of $P_R(HE)$ and $P_R(IHC)$ is applied to the WSIs from those two separate modalities. However, because of the lack of interdependency between the $P_R(HE)$ and $P_R(IHC)$, a correlation mapping function is required to ensure that any arbitrary $p_{ij}^{HE}$ detected as having malignancy, should be mirrored back into $P_R(IHC)$ to obtain a corresponding $p_{ij}^{IHC}$. In equation (3), we show the inter-modality patch mapping, where $I(.)$ is denotes the inter-modality image mapping function.

$$I : P_R(HE) \Longrightarrow P_R(IHC) \quad (3)$$

Also given $\forall\ p_{ij}^{HE} \in P_R(HE)$ and $\forall\ p_{ij}^{IHC} \in P_R(IHC)$, we will conclude that equations (4) is satisfied when the mapping function $I(.)$ holds

$$I(p_{ij}^{HE}) = p_{ij}^{IHC} \bigwedge I(p_{ik}^{HE}) = I(p_{kj}^{IHC}) \qquad (4)$$

The requirement that must be satisfied includes surjective and injective conditions which in turn makes $I(.)$ a bijective function. The injective component of the equation ensures that there are no two elements of $P_R(HE)$ mapping to the same output in $P_R(IHC)$. The backward injective mapping is also enforced by the forward operation. This implies that should the output of any two tiles in $P_R(HE)$ be equal on $I(.)$, then the original inputs must be equal. Similarly, the surjective component of the equation ensures that there is at least one match of $P_R(IHC)$ for every item in $P_R(HE)$.

The inter-modality mapping function also allows for ensuring that every $p_{ij}^{HE}$ in $P_R(IHC)$ should be able to merge to form a complete WSI for H&E as denoted by $S_{HE}$. Similarly, it is desirable that tiles extracted from $S_{IHC}$ (representative of WSI of IHC) into $p_{ij}^{IHC} \in P_R(IHC)$, should be composable into a single WSI after operations described in sections 3.3 – 3.5. This implies that both $P_R(IHC)$ and $P_R(IHC)$ may have $\widetilde{p_{ij}^{HE}}$ and $\widetilde{p_{ij}^{IHC}}$ contained in them respectively, where the $\tilde{a}$ represents a situation of detected malignancy or some HER2 positive status detected in both H&E and IHC WSIs. These detection and status scoring procedures are described in the following subsections.

### 3.3 Patch-wise score assignment strategy

In this subsection, the focus is the derivation of HER2 score, that is and $\widetilde{p_{ij}^{IHC}}$, from patches $p_{ij}^{IHC}$. To achieve this, first a transformer-like model ($\mathcal{L}$) for pixel-level annotation applied to the $p_{ij}^{IHC}$ to obtain $x$ representing pixel values for dimension $WxH$ (e.g 512x512). In equation (5), the pixel values for each tile or patch are computed, while equation (6) shows how frequency of occurrence for a 4-way HER2 scoring is derived. Note that $\mathcal{L}$ (see subsection 3.5) is trained to yield pixel values for each category in the 4-way {0, 1+, 2+, and 3+} HER2 status corresponding to {1, 2, 3, and 4} otherwise, it assigns a 0.

$$x = \mathcal{L}(p_{ij}^{IHC}) \qquad (5)$$

$$Nx = \left| \sum_{u=0}^{H} \sum_{v=o}^{W} x_{u,v} \right|, \qquad \forall x_{u,v} \in \{1, 2, 3, 4\} \qquad (6)$$

The notation $Nx$ infers the number of pixels with any of the categorization in the 4-way status scoring. Meanwhile, if we are simply interested in binary scoring, that is 0 or 1+ (HER2-negative), 2+ (equivocal/borderline), and 3+ (HER2-positive), then the equation will be constrained by $\forall x_{u,v} \in \{1, 4\}$. Also, to find the pixel distribution for a particular categorization, we constrain the equation by any of the $Nx_{\{1\}}, Nx_{\{2\}}, Nx_{\{3\}}$ and $Nx_{\{4\}}$. Pathologists are often interested in percentage coverage of score on a tile of WSI. Hence, this patch-wise percentage computation follows: $\frac{Nx*100}{H*W}$. However, since the focus of this study is the 4-way categorization, patch-wise HER2 status scoring ($\mathcal{T}^{p_{ij}^{IHC}}$) follows in equation (7):

$$\mathcal{T}^{p_{ij}^{IHC}} = \max(Nx_{\{1\}}, Nx_{\{2\}}, Nx_{\{3\}}, Nx_{\{4\}}) \oplus \mathcal{C}(p_{ij}^{HE}) \oplus \mathcal{M}(p_{ij}^{IHC}) \qquad (7)$$

Note that both $\mathcal{C}(p_{ij}^{HE})$ and $\mathcal{M}(p_{ij}^{IHC})$ are classifiers (see subsection 3.4) acting on H&E and IHC WSIs, respectively. As we are more interested in scoring WSIs rather than patch-wise scoring, we aggregate these first into blocks and then WSIs. Since we are interested in HER2 status scores 2+ and 3+, IHC WSIs single score computation follows in equation (8), while the computation for the percentage of region covered by HER2-positive is described by equation (9).

$$WSI_{score} = \max\left(\mathcal{T}^{p_{ij}^{IHC}}\right), \quad \forall \mathcal{T}^{p_{ij}^{IHC}} \in \{Nx_{\{3\}}, Nx_{\{4\}}\} \qquad (8)$$

$$WSI_\% = \frac{\left|\sum \mathcal{T}^{p_{ij}^{IHC}}\right| * 100}{H * W}, \qquad \forall \mathcal{T}^{p_{ij}^{IHC}} \in \{Nx_{\{3\}}, Nx_{\{4\}}\} \qquad (9)$$

The pseudocode describing this HER2 status scoring strategy is encoded in Algorithm 1. The algorithms show that patches of both H&E and IHC are passed as input, while the output is the WSI single score and the percentage of coverage are the output. Lines 1 – 3 of the algorithms show parameter initialization. The main scoring strategy for patch-wise and WSI-wide HER2 status scoring are encoded on lines 4 – 10, and these are consistent with the equations explained in previous paragraphs.

---

**Algorithm 1** HER2 patch-wise to WSI status scoring strategy

**Input**: $P_R(HE)$ and $P_R(IHC)$, $\mathcal{L}$, $\mathcal{M}$, $\mathcal{C}$
**Output**: Whole slide level HER2 score and percentage of abnormality coverage

1     $WSI_\% = 0$
2     $WSI_{score} = \emptyset$
3     $scores = \{\}$
4     **for** $i \leftarrow 1$ **to** $H$ **do**
5        **for** $j \leftarrow 1$ **to** $W$ **do**
6           $p_{ij}^{HE} = P_R(HE)$ and $p_{ij}^{IHC} = P_R(IHC)$
7           $x, \grave{x}_1, \grave{x}_2 = \mathcal{L}(p_{ij}^{IHC}), (\mathcal{M}(p_{ij}^{IHC}), \mathcal{C}(p_{ij}^{HE})$
8           $\mathcal{T}^{p_{ij}^{IHC}} = x \oplus \grave{x}_1 \oplus \grave{x}_2$
9           $scores \leftarrow \mathcal{T}^{p_{ij}^{IHC}}$
10    $WSI_\% = \frac{(scores)\%}{W.H}, (scores)$
11    return $WSI_\%, WSI_{score}$

---

In the following subsections, a detailed description of all the transformer-based architectures $\mathcal{L}$, $\mathcal{M}$, and $\mathcal{C}$, are presented. We highlight their application to the classification and scoring process in the end-to-end pipeline.

### 3.4 Tumor and stain intensity classifier

Tumor or malignant tissue classification with H&E WSIs is critical to the end-to-end pipeline proposed in this study. Similarly, accurately predicting the intensity of stain in IHC WSIs supports the HER2 status scoring process. In Figure 2, interaction among the transformer-based architectures, namely stain intensity vision transformer ($\mathcal{L}$), Segmentation map prediction transformer ($\mathcal{M}$), and H&E hierarchical transformer ($\mathcal{C}$) are illustrated. First, the patch generator accepts WSIs from the two modalities and outputs patches for embedding computation into the various transformers. Output from the $\mathcal{C}$ model returns heat-mapped classified patches which are further imploded into an annotated H&E WSIs. On the other hand, both $\mathcal{L}$ and $\mathcal{M}$ and accept patches from IHC WSIs and apply stain intensity classifier and segmentation map prediction operations on the patches.

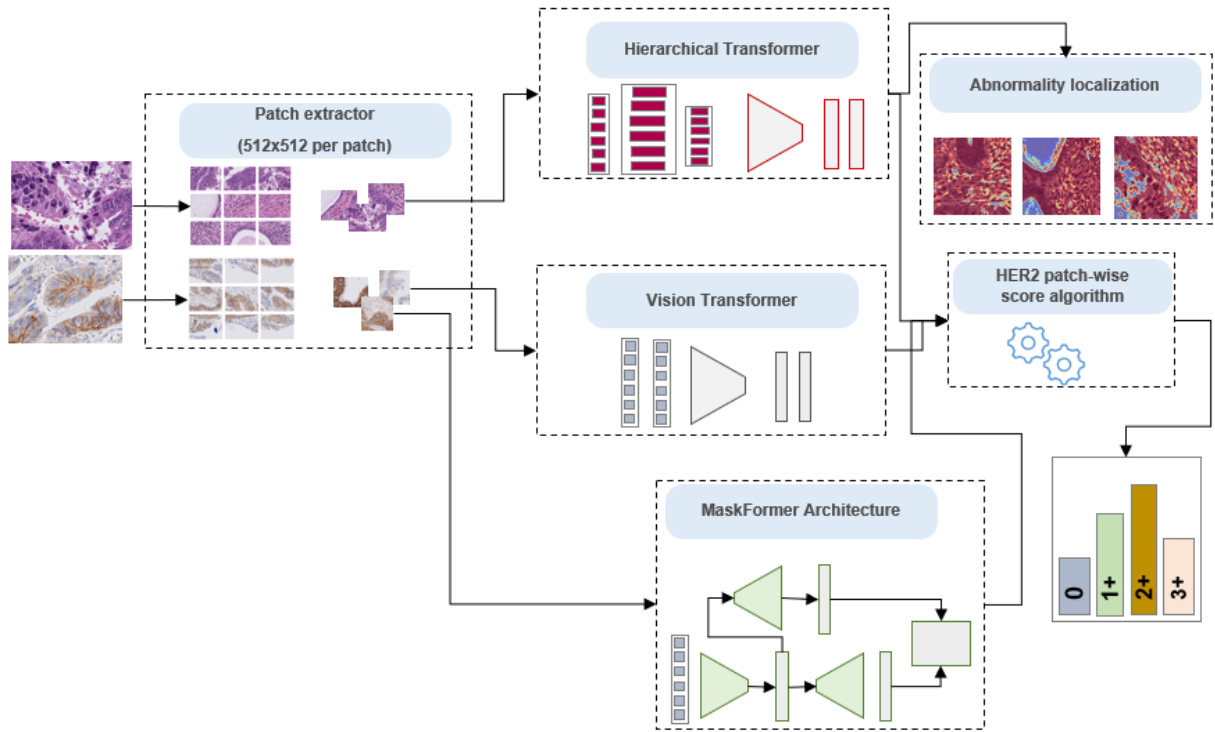

**Figure 2**: An illustration of the interaction among transformer-based architectures $\mathcal{L}$, $\mathcal{M}$, and $\mathcal{C}$ for HER2 status scoring

The output from the transformer-based $\mathcal{L}$, $\mathcal{M}$, and $\mathcal{C}$ models are combined to compute the patch-wise HER2 status score. As previously described in Algorithm 1, the patch-level scores are aggregated into a WSI-level score. Hence, the reason the figure shows how the algorithm returns the final output, which is the HER2 scores.

### 3.5 MaskFormer for HER2 status intensity region segmentation

The application of $\mathcal{M}$ as described in the previous subsection shows that segmentation maps are generated to provide pixel-level categorization which produces a mask classification corresponding to each HER2 status (0, 1+, 2+, and 3+). The implication is that output from the model will be a binary image per patch/tile, indicating which category a pixel belongs. So, rather than strictly classifying pixels, unique masks are predicted to represent instances of HER2 status categories. In this study, MaskFormer [48] architecture is leveraged to derive the semantic segmentation (a classification of the IHC patches into regions of different unique HER2 status) and instance segmentation (same as the localization of HER2 status on IHC patches using instance masks). Hence, it is the aim of this study to fine-tune $\mathcal{M}$ so that accurately predicts a combination of semantic segmentation and instance segmentation through per-pixel mask detection. This is made possible because MaskFormer has four distinct modules namely the backbone (for feature learning and extraction using neural network), transformer decoder (capable of generating segmentation of all regions in each IHC patch through the features extracted by backbone module), pixel decoder (returns pixel-level embeddings using the features generated by the backbone module), and class and mask prediction (uses multi-layer perceptron for classification of regions in IHC patch into all the 4-way HER2 statuses). Note that the class and mask prediction module in the MaskFormer also generates binary masks for every region or segment in the IHC patch.

In summary, this section provided a detailed description of the end-to-end pipeline for HER status scoring. We highlighted the inter-modality image mapping function and the strategy for scoring HER2 patches/WSIS from image features, which was illustrated using an algorithm. Finally, the system of transformer-based architecture used for feature learning and segmentation maps was also discussed. In the next section, computational environment setup and experimentation are discussed.

### 4. Computational Settings and Image Dataset

To promote reproducibility of the methodology proposed in this study, a detailed description of the parameter settings for all models trained for the end-to-end pipeline used for experimentation is provided in this section. Also, computational resources mirroring the experimental environment are outlined to support verification of the results and setting up for reproducibility. This is followed by the pathologist's description of the dataset curation and annotation process. Also, detailed statistics of the two modalities of datasets used are presented.

### 4.1 Model parameter settings and computational environment

The end-to-end pipeline consists of three major vision transformer-based architecture $\mathcal{L}$, $\mathcal{M}$, and $\mathcal{C}$ C, fine-tuned into models. Table 1 outlines the choice of hyperparameters applied for the training of each of the models.

**Table 1**: Summary of parameter configurations for transformer-based models in the end-to-end pipeline

| Model | Hyperparameters | Input Size | Data partition | Epoch |
|---|---|---|---|---|
| Stain intensity ($\mathcal{L}$) | Learning rate: 2e-4 Batch size:16 | 512x512 | 10627 training, 1782 evaluation, and 94 testing. | 20 epochs |
| Segmentation map ($\mathcal{M}$) | Learning rate: 5e-5 Batch size:2 | 512x512 | 1055 training, 56 evaluation, and 278 testing | 50 epochs |
| Tumor classifier ($\mathcal{C}$) | Learning rate: 2e-4 Batch size:16 | 244x244 | 14172 training, 2375 evaluation, and 126 testing | 20 epochs |

Google Colab GPUs were explored to ensure sufficient computing power is used for the training. Specifically, the Nvidia A100 GPU was applied for training. All three $\mathcal{L}$, $\mathcal{M}$, and $\mathcal{C}$ C models were separately trained using the A100 Tensor Core GPU with a memory size of 80 GB. Implementation was done using the Python 3 programming language alongside other dependencies, with a mix of Pytoch and TensorFlow configurations. Meanwhile, testing of the trained model was implemented on a personal computer (PC) with a RAM size of 32GB and a CPU configuration of Intel Core i7 with a processing speed of 1.4GHz on the Windows 11 operating system.

## 4.2 Image Datasets

Two modality datasets of H&E and IHC WSIs were curated for the experimentation of the proposed method. The biomedical scientist uses a dataset of 13 cases of p53-abnormal endometrial carcinoma tissue to train this model with both H and E and HER2 IHC (4B5 Clone). Among these cases were various sample types, including uterine resections and endometrial biopsies, to help add variation, mimicking clinical practice. These images were then segmented into fixed-size patches measuring 143 x 143 microns. The H & E patches will then be extracted and annotated manually to identify relevant cellular features, including tumor cells, blood cells, normal stroma, and common tissue artifacts. These annotations will facilitate the model to highlight areas of tumor tissue to accurately predict the tumor percentage of the WSI. This is a necessary precursor used in the eventual HER2 score. Training on HER2 IHC WSIs will be conducted in a similar way, with the annotation of patches; however, the classification on these images is more complex. Tumor heterogeneity and scorer subjectivity play a role in making this a more difficult task. HER2 expression is classified as either 0, 1+, 2+, or 3+. The criteria for classification include staining intensity and pattern, which are then analysed alongside the H and E tumor percentage to give an overall WSI score according to [49]. Accurate patch annotations are important as tissue artifacts, tissue folds, or background staining can all mask or mimic positive signals, which in turn can affect the overall accuracy of the patient score.

In Figure 3, we capture samples of the two modalities of images used for the experiment, and for each H&E patch, a corresponding IHC-stained patch was extracted, and the pathologist observed the abnormality label listed in (a-d). The H&E annotation shows if the patch has a tumor or not, while the IHC patches show the HER2 score status inferred by the pathologist. These datasets were used for training the stain intensity vision transformer ($\mathcal{L}$) and H&E hierarchical transformer ($\mathcal{C}$).

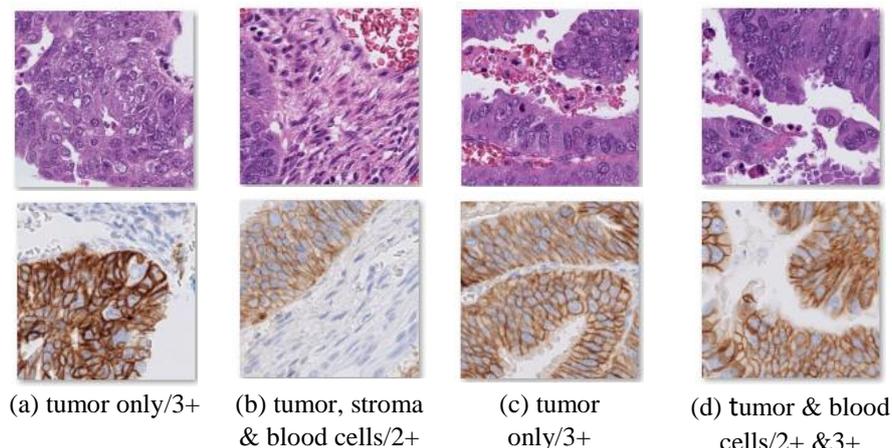

(a) tumor only/3+  (b) tumor, stroma & blood cells/2+  (c) tumor only/3+  (d) tumor & blood cells/2+ &3+

**Figure 3**: Samples of H&E and HER2 from the private dataset collected from the hospital, with each H&E-HER2 pair showing corresponding labels

To prepare a dataset for the pixel-level HER2 status region segmentation, we adapted a super-pixel trained model [50] which can divide every IHC patch into coherent regions. The model automatically detects separating regions in every patch, and also provides human intervention to improve the granularity of the super-pixel operation through mouse-guided annotation. The transformer and machine learning-assisted models helped in preparing datasets, which achieved accurate segmentation of patches supported by annotation/label (segmentation masks) to train our segmentation map prediction transformer ($\mathcal{M}$). In Figure 4, we show the segmentation masks for the four HER2 statuses (0, 1+, 2+, and 4+) and a background (plain). For each IHC patch, a corresponding segmentation mask is displayed as seen in Figure 4(a). In Figure(b), a graph showing pixel-value distributions across the four categories and the background (plain) is included. Figure 5 shows the outcome of overlaying the segmentation masks on the original IHC patch to help pathologists easily spot the difference across regions.

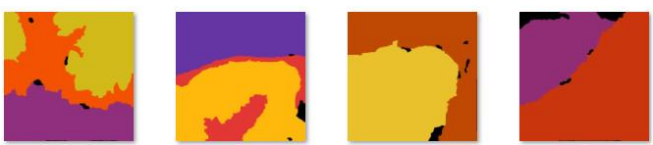

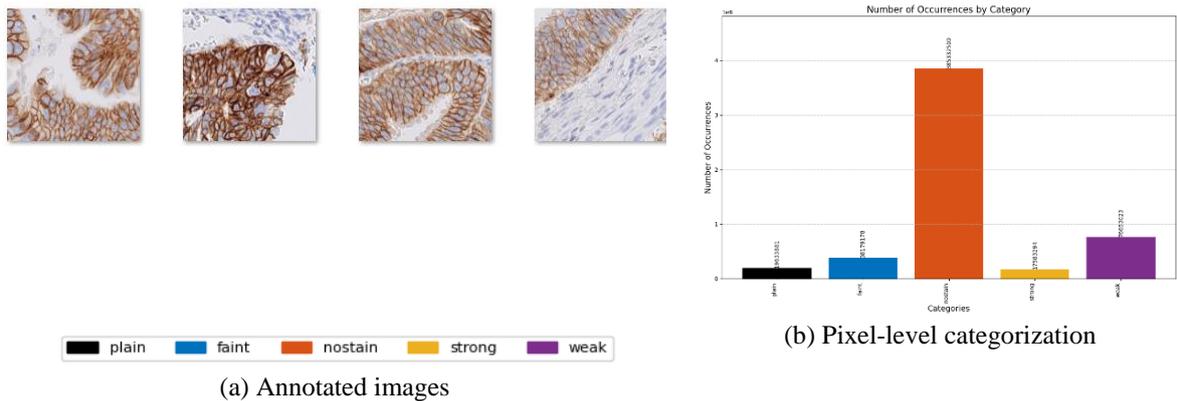

(b) Pixel-level categorization

(a) Annotated images

**Figure 4**: Some selected samples of (a) HER2 IHC (immunohistochemistry) with corresponding segmentation bitmap annotated using [50], and (b) graphing of the pixel-level categorization in samples

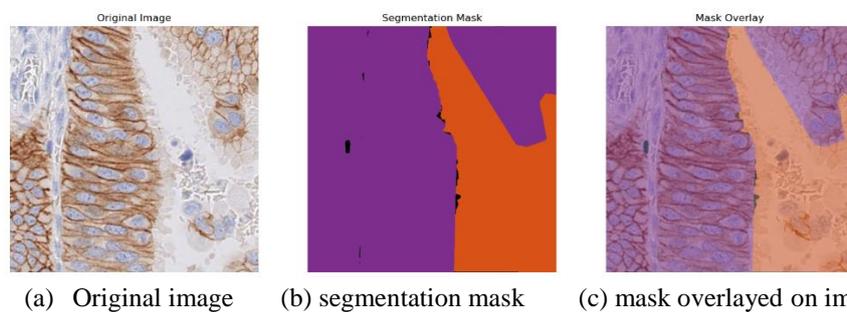

(a) Original image     (b) segmentation mask     (c) mask overlayed on image

**Figure 5**: A single HER2 with (a) weak staining intensity and an overlay of the (b) segmentation mask, which shows (c) a combination of segmentation bitmap, annotations, and image

Considering the detailed experimental setup and the parameter configuration for the three major models sitting in the end-to-end pipeline, the complete implementation was completed. Moreover, the pathologist and AI-driven dataset curation and annotation allowed for applying the pipeline to experimentation. In the following section, a detailed explanation of the results obtained from the experimentation described in this section is presented.

**5. Results and Discussion**

In this section, the performance evaluation of the proposed model is investigated and reported. Results gathered during the experimental setup described in the previous paragraph are outlined here. First, we conducted a performance study of the end-to-end pipeline's usability. This was followed by an investigation to examine the applicability of the pipeline to fully process WSIs of H&E and IHC for HER2 status scoring. Moreover, the tumor detection capability of the pipeline was also evaluated. We supported the evaluation of the proposed method with feedback and validation from human experts (pathologists). Furthermore, a comparative analysis with some state-of-the-art (SOTA) studies is also carried out to contextualize the proposed method in related works.

**5.1 Evaluation of the Base Models**

The three separate models described in the proposed pipeline are first evaluated to test their suitability for the task of scoring IHC images. In this subsection, the H&E image tumor classification models, the IHC image stain intensity classification models, and the stain-segmentation model are evaluated. Two separate models, namely the vit-small-patch16-224 and vit-base-patch16-224, were trained to address the tumor identification task. The same models were applied to the stain-intensity classification problem. In Table 2, we outline the results obtained based on the classification accuracy, precision, F1-score, recall, and sensitivity metrics. The vit-small was trained for 20 epochs on a 4GB GPU, and the result shows a very impressive performance, whereas the vit-based was trained for 5 epochs on high high-GPU Google Colab platform. Results show that the performance of the vit-based is not comparable with that of the vit-small. Considering this superiority in performance, the vit-small model was selected for the tumor classification task in H&E images.

**Table 2**: Classification performance evaluation of the H&E and HER2 models (vit-small-patch16-224 and vit-base-patch16-224) based on metrics: precision, F1-score, sensitivity, recall, and accuracy

| Task | Model | Precision | F1 | Sensitivity | Recall | Accuracy |
|------|-------|-----------|-----|-------------|--------|----------|
| H&E classification | vit-small-patch16-224 | 1.0 | 1.0 | 1.0 | 1.0 | 1.0 |
| | vit-base -patch16-224 | 0.69 | 0.339 | 0.531 | 0.531 | 0.4126 |
| HER2 Classification | vit-small -patch16-224 | 0.875 | 0.903 | 0.933 | 0.933 | 0.9148 |
| | vit-base -patch16-224 | 0.982 | 0.991 | 1.0 | 1.0 | 0.9400 |

On the other hand, stain-intensity classification on IHC/HER2 images returned high performance on both the vit-small and vit-based models. These two models were trained for the duration of 20 epochs on a 4GB GPU computer. Results show that the vit-based model outperformed the vit-small. A confirmation that when the vit-based used on the H&E samples is trained further, we will obtain better performance. Meanwhile, given the classification accuracy of 0.9148 and 0.940 on the vit-small and vit-based models, both models were therefore applied to the stain-intensity classification task. The outcome of the application of the tumor and stain-intensity classification models is further discussed in the following paragraphs.

The confusion matrix outputs and receiver operating characteristic curve (ROC) graphs for all the models trained for both classification tasks are shown in Figure 6. Figures display under Figure (a-c) show the confusion matrix plots for vit-small and vit-based, and a ROC curve combining their true positive rate (TPR) versus false positive rate (FPR). While the confusion matrices show a very impressive performance for vit-small, and also somewhat for vit-based, the ROC curve clearly shows that vit-based underperformed in differentiating TPR from FPR. Again, this outcome motivated the choice of the vit-small model for further application to the task described in the next subsection.

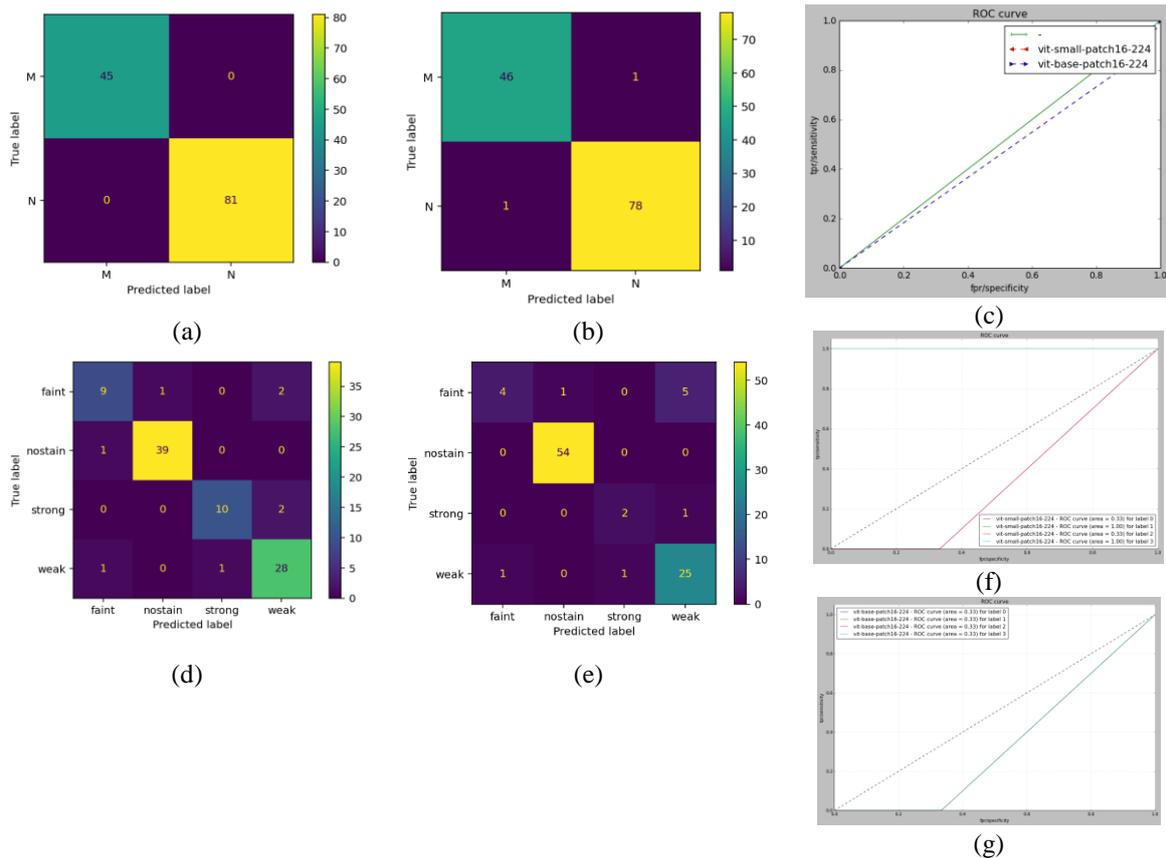

**Figure 6:** Performance evaluation of the (a) H&E samples on vit-small-patch16-224, (b) using vit-base-patch16-224 (c) ROC curve showing performance using vit-small and vit-based on H&E samples. (d) HER2 samples on vit-small-patch16-224, and (e) HER2 samples on vit-base-patch16-224, and (f) ROC curve showing performance using vit-small and vit-based on HER2 samples.

In the same figure, particularly, those labeled Figure 6 (d-g), performance for the stain-intensity classification models is displayed for the confusion matrix and ROC curve. Interestingly, we observed that the confusion matrix for both vit-small and vit-based shows good patterns. Meanwhile, the ROC curve, separated for each of the four classes (no-stain, faint-stain, weak-stain, and strong-stain), shows that these models performed well specifically in correctly classifying stain in some classes compared to the other.

The stain segmentation model was evaluated using the Intersection over Union (IoU) metric. The IoU metric has proven to be very useful in evaluating the accuracy of how well the segmentation model detects regions of interest within the whole image space. This detection by the model is compared with the ground truth by computing the size of the intersection between the area of the region detected with what is reported in the ground truth. In the study, the mean of IoU was calculated by finding the average of IoUs. By taking this approach, our study is able to quantify the performance and capability of the model in selecting each pixel in the IHC images. The outcome of this demonstrates how the model helps to segment regions (or related pixels) in the IHC image according to their stain intensity. In Figure 7, we plot the performance of the segmentation model after training it for 5 epochs and 50 epochs. It is always desirable that the mean IoU range between 0.0 – 1.0, with values closer to 1.0 evidence of good performance in the correct classification of pixels in the images. In graph for the 5 epochs shows that while at epoch-2 the model attained a 0.7 mean IoU, this performance fell to about 0.52 at epoch-4. Unsatisfied with this performance, we decided to train the model further to investigate the possibility of improvement. The result obtained for the 50-epoch training shows that the mean IoU improved and rose to 0.9, though with some evidence of instability around that peak.

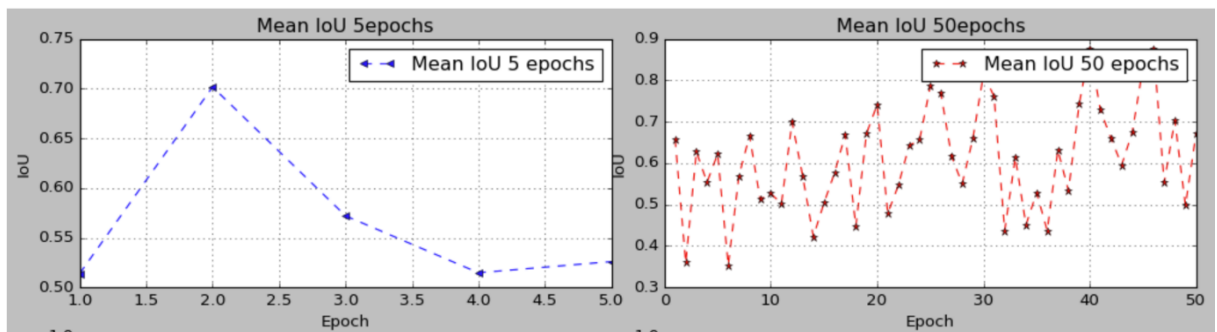

**Figure 7:** Performance evaluation of the SegFormer model showing the mean IoU values obtained during the training phase over (a) 5 epochs and (b) 50 epochs, respectively

The validation mean of IoU for the model trained for 50 epochs returned a value of 0.6719, and the validation mean IoU value of 0.5260 for the model trained for 5 epochs. Considering that the performance of the first model is better than the latter, we selected this to address the problem described in the next subsection.

Furthermore, we evaluated the applicability of the classification models to understand how suitable they are for clinical utility and net benefit. To achieve this, decision curve analysis (DCA) and the associated graphs were applied to check this suitability. The importance of the DCA is to ensure that all the models are tested beyond the previous metrics (accuracy, precision, F1-score, recall, sensitivity, TPR, and FPR) discussed above. The evaluation strategy of the DCA is mainly to ensure that a model's performance sits above the 'treat none' (showing the net benefit when no one is treated) and 'treat all' (showing the net benefit when all are treated) curves. It is often desirable that the net benefit of the model's performance lies above those of 'treat none' and 'treat all', and this net benefit highlights the variation of the benefits of true positives (accurately identified high-risk cases that need to be treated) and the harms of false positives (accurately identified cases with low risk). While the y-axis often shows the computed net benefit, the x-axis calibrates using the likely threshold probabilities, indicative of high-risk cases.

In this study, we chose the maximum probability threshold of 0.30 to calibrate the x-axis, given the clinical understanding that the threshold is suitable enough to suspect a malignant case and stain intensity on H&E and IHC images, respectively. In Figures 8(a-b) and 8(c – d), the DCA graph plots for the malignant case and stain intensity on H&E and IHC images, respectively, are shown. It is often desirable that the prediction model curve lies above the 'treat none' and 'treat all' though the 'treat none' may intercept at point 0 of the y-axis. Given a 0 – 30 probability range of risk thresholds, malignant DCA graphs on vit-small show a strong overlap with the 'treat none' and 'treat all' lines, while the vit-based lies above the 'treat none' and 'treat all' lines. This therefore indicates that the vit-based shows more clinical benefit over the vit-small. Similarly, the multiclass classification nature of the stain-intensity model was unbundled into a 'one vs all' binary form to plot the DCA graph. The results obtained show that the vit-small appears to accurately identify cases of weak stain, while the vit-based appears to accurately identify the faint-stain and strong-stain. This implies that both models have some measure of clinical relevance in the classification of the intensity of stain in IHC samples, leading to HER2 scoring.

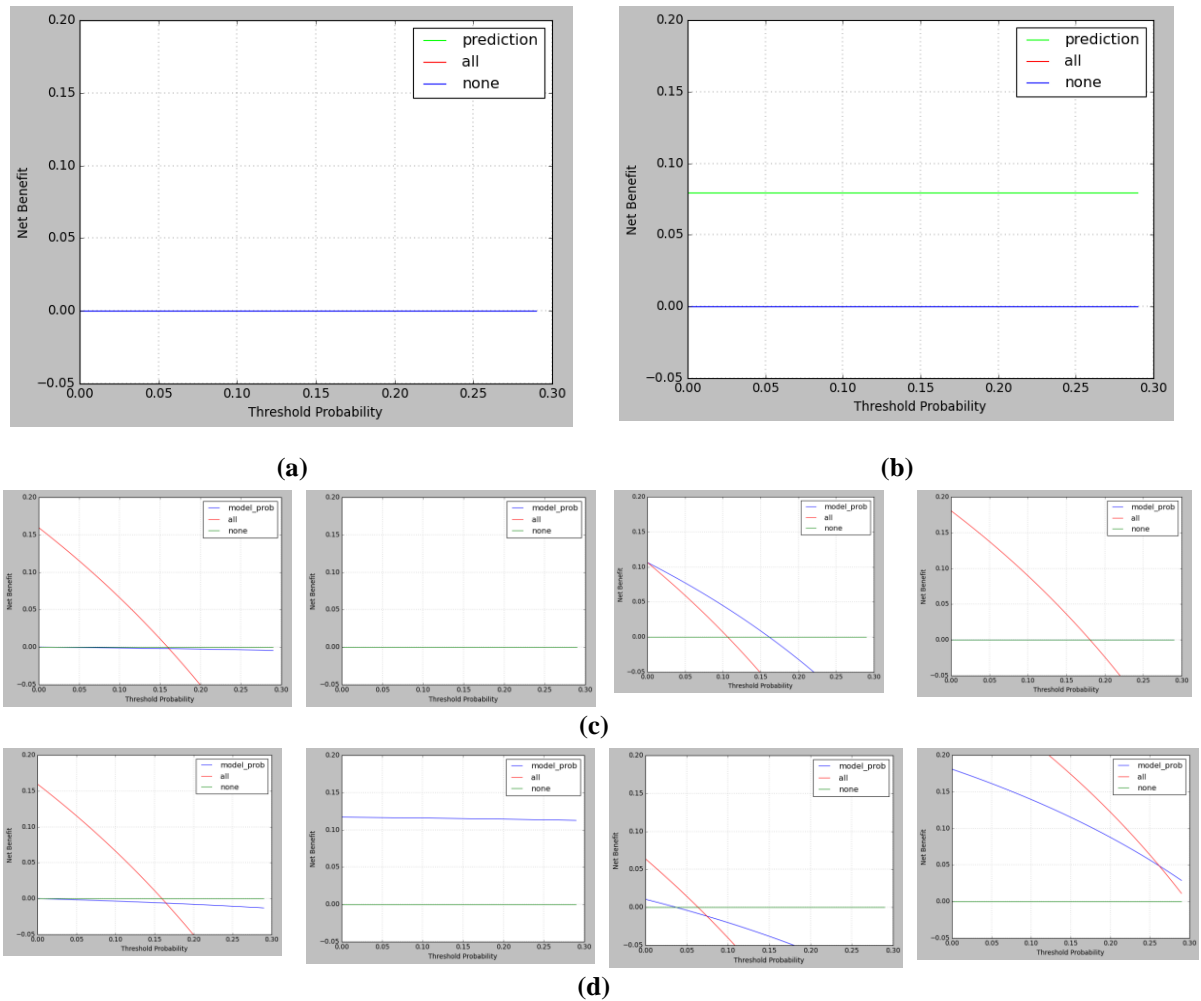

**Figure 8:** Decision curve analysis (DCA) of models on the test set (a) H&E with vit-small-patch16-224, (b) H&E with vit-base-patch16-224, (c) HER2 with vit-small-patch16-224 separated on faint, no-stain, strong and weak classes respectively, and (d) HER2 with vit-base-patch16-224 separated on faint, no-stain, strong and weak classes respectively

In summary, the DCA graph reveals that the malignant classification model on H&E is clinically beneficial at a risk threshold of 0.30 probability. Similarly, the observation shows that the stain-intensity classification model is clinically beneficial in supporting the process of scoring HER2 expression in IHC images. In the next subsection, we shall be looking over the impacts of applying these models to the whole pipeline proposed in this study.

### 5.2 HER2 Scoring evaluation

The main aim of this study is to score the HER2 expression in IHC images and WSI and to jointly classify the stain intensity in these images. In addition to these core tasks, the study also aims at detecting tumors in patches of WSIs and at the WSI level. In this subsection, we demonstrate the outcome of the study based on the strategic aim. To confirm the satisfiability of this aim, we shall be investigating the performance of the pipeline on HER2 scoring and tumor detection on a patch-wise, region of interest (ROIs), and also WSI levels.

We have selected patches from the WSI slide collected from case number 6 (c6). To facilitate the path-wise processing, ROIs were extracted from the full WSIs in a block-wise allocation. Figure 9 shows the pairs of image patches extracted from case 6 region 4 (C6R4) and region 6 (C6R6). Each row associated with the IHC/HER2 score shows three patches of IHC images of the original, overlayed with a predicted segmentation map, and annotated with stain-intensity classification with HER2 score. The row labeled 'H&E with no tumor' shows the outcome of the malignant classifier with the original image patch alongside the heatmap and annotated label. Obviously, the figure shows that the HER2 score for the IHC image is 0, and clear indication that there is no evidence of IHC stain detected. also, the second row on the figure shows that the H&E patch is malignant and does have a tumor, as the annotated label shows the 'N' for 'no-stain' patch. Most of the heatmap portion of the H&E simply highlights the presence of stroma, gland, and tumor.

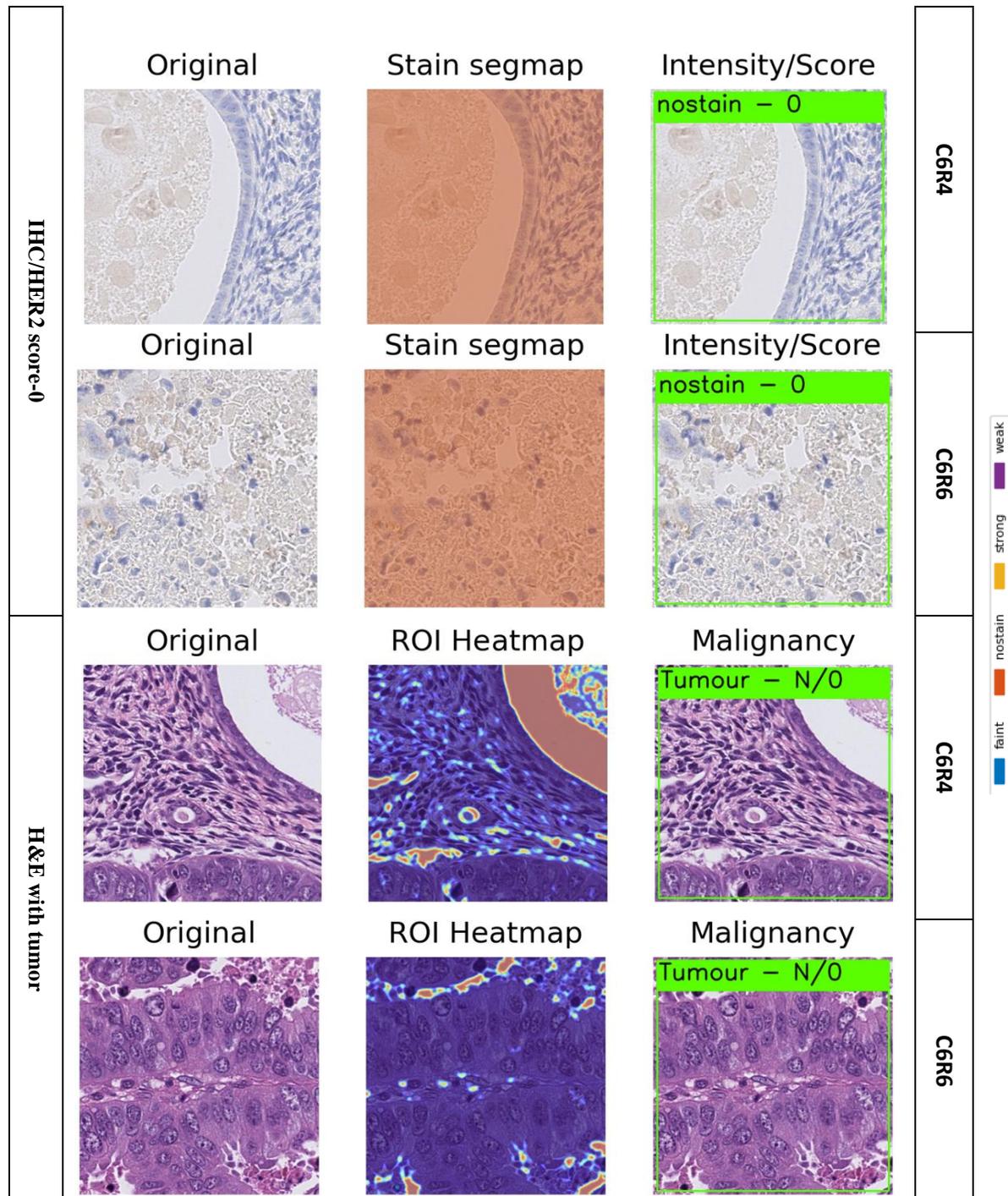

**Figure 9**: The patches of IHC with HER2-score 0 (no-stain) and H&E with malignant (stroma, gland, and tumor), extracted from WSI-case-6, showing segmentation annotation and attention heatmaps

Furthermore, HER2 scoring for 1+ was also evaluated using the proposed pipeline, and the results obtained are shown in Figure 10. The upper row of the figure captures two lines of IHC images extracted from C6R4 and

C6R6, respectively. Interestingly, we see a very accurate scoring being achieved by our model with the two lines of IHC showing a 1+ HER2 score for the C6R4 and C6R6 samples. Similarly, the two lines also show the overlay of the stain map segmentation to illustrate a section of the patch and the various stain intensities. Using the color-map key, it is obvious that the patch from C6R4 contains portions of faint-stain and while the patch from C6R6 indicates the presence of both weak-stain and faint-stain, and our model accurately predicted a 1+ HER2 score. Correspondingly, the lower row of the figure shows the H&E patches correctly classified as having a tumor, and the heatmap applied to those malignant patches. Interestingly, the annotation correctly shows 'NF/1+' and 'NFW/1+' labels for C6R4 and C6R6, respectively.

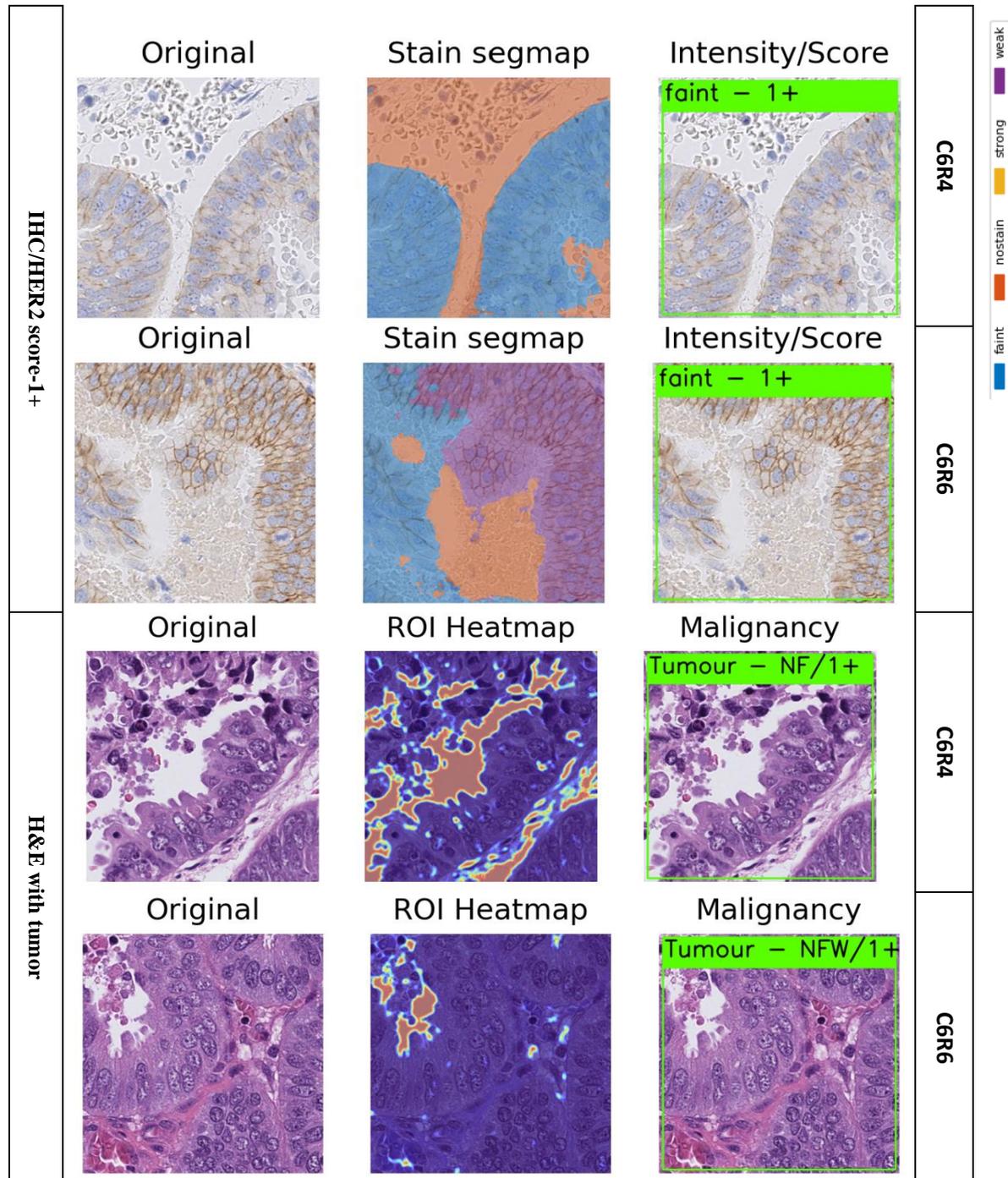

**Figure 10:** The patches of IHC with HER2-score 1+ (faint stain) and H&E with malignant (tumor), extracted from WSI-case-6, showing segmentation annotation and attention heatmaps

The evaluation of the pipeline on the HER2 scoring for 2+ was applied, and the result obtained is shown in the first row in Figure 11. This first row captures the scoring and segmentation of patches drawn from C6R4 and C6R6. It is clear that the IHC images on line one of row one have a combination of faint and weak staining, as the stain segmentation mapping shows; however, while the stain intensity was correctly classified, the HER2-score shows that it is 0. Similarly, the second line of the row also shows an annotation of the staining segmentation map of weak and strong stain. The stain-intensity classifier correctly classified the stain as weak, but wrongly classified the HER2 scoring as 0. This indicates the slight bumps in the scoring of the pipeline, though most of the cases have been successfully and accurately classified. On the other hand, the second row of the figure shows the H&E patch classification for C6R4 and C6R6 on lines 1 and 2, respectively. Interestingly, the heatmap and the malignant classification for both C6R4 and C6R6 patches were correctly performed. In addition, the annotation for staining correctly captures all the detected stain intensity as represented by the 'NFW', which denotes 'no-stain, faint-stain, and weak-stain'.

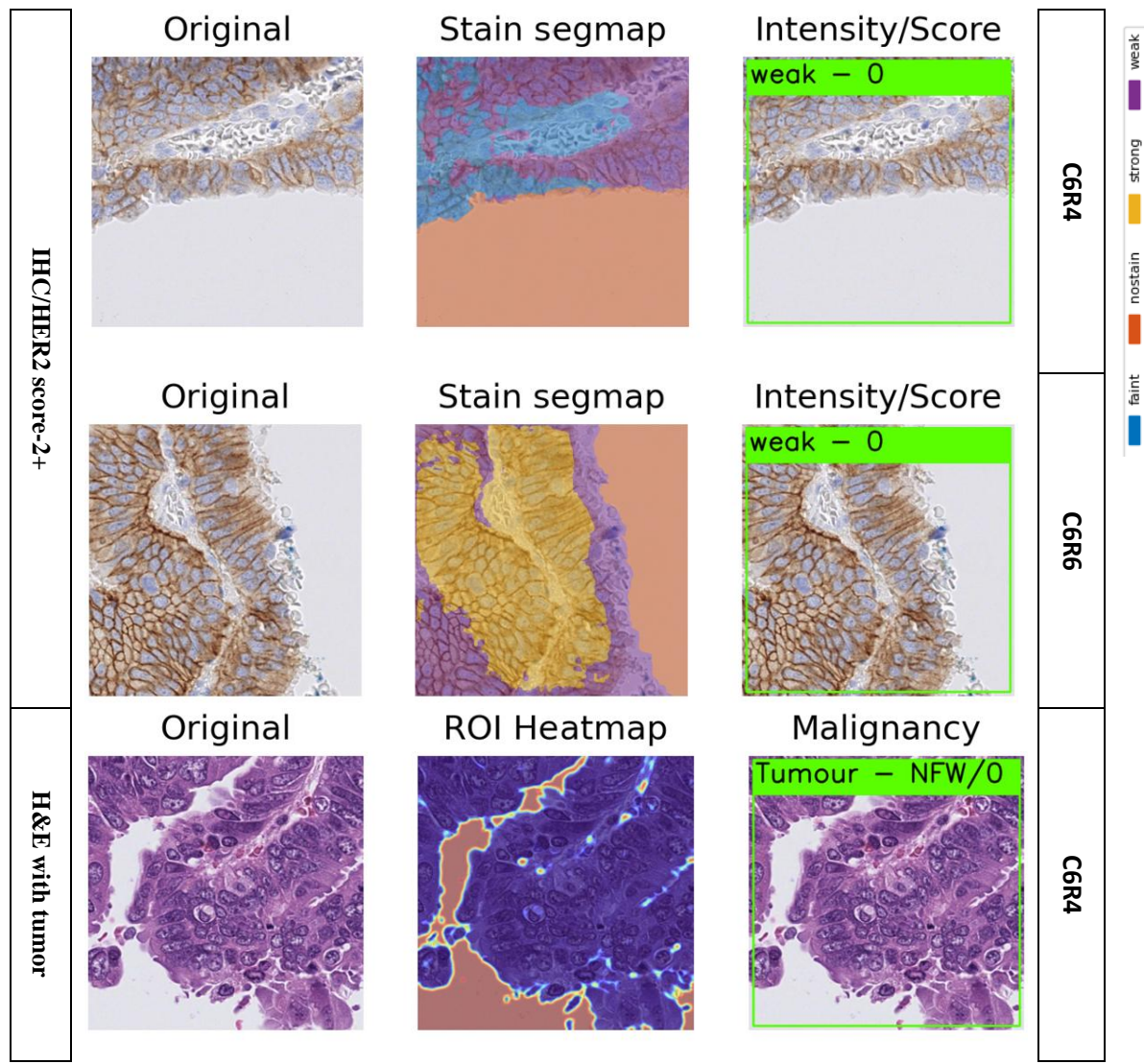

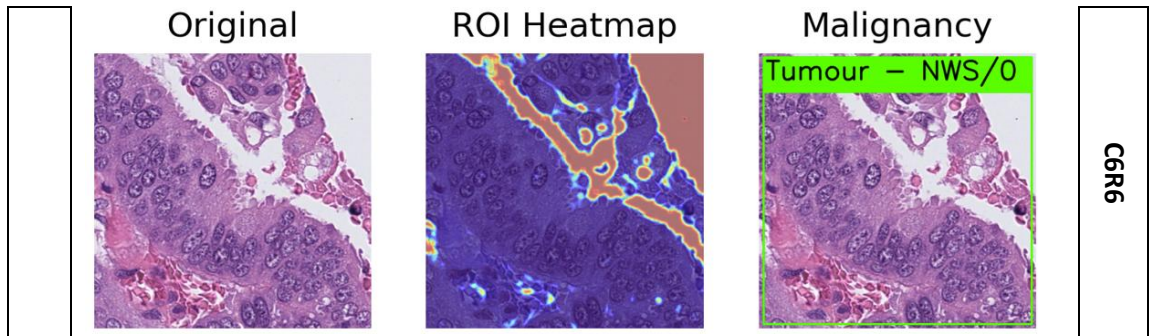

**Figure 11:** The patches of IHC with HER2-score 2+ (weak stain) and H&E with malignant (tumor), extracted from WSI-case-6, showing segmentation annotation and attention heatmaps

Finally, the HER2 score for 3+ was also evaluated on the patch-wise level using samples drawn from the C6R4 and C6R6 blocks. In row 1 of Figure 12, the first line shows a very impressive segmentation of the stain mapping for strong and weak stains. A similar mapping and segmentation was observed on the second line of the first row. HER2 scoring of both IHC patches extracted from C6R4 and C6R6 was accurately scored as 3+. On the second row, where the H&E patches for C6R4 and C6R6 were captured, it is clear from the heatmap and annotated patches that the malignancy was correctly detected. Furthermore, these were also correctly annotated with the stain intensity seen in the IHC patches.

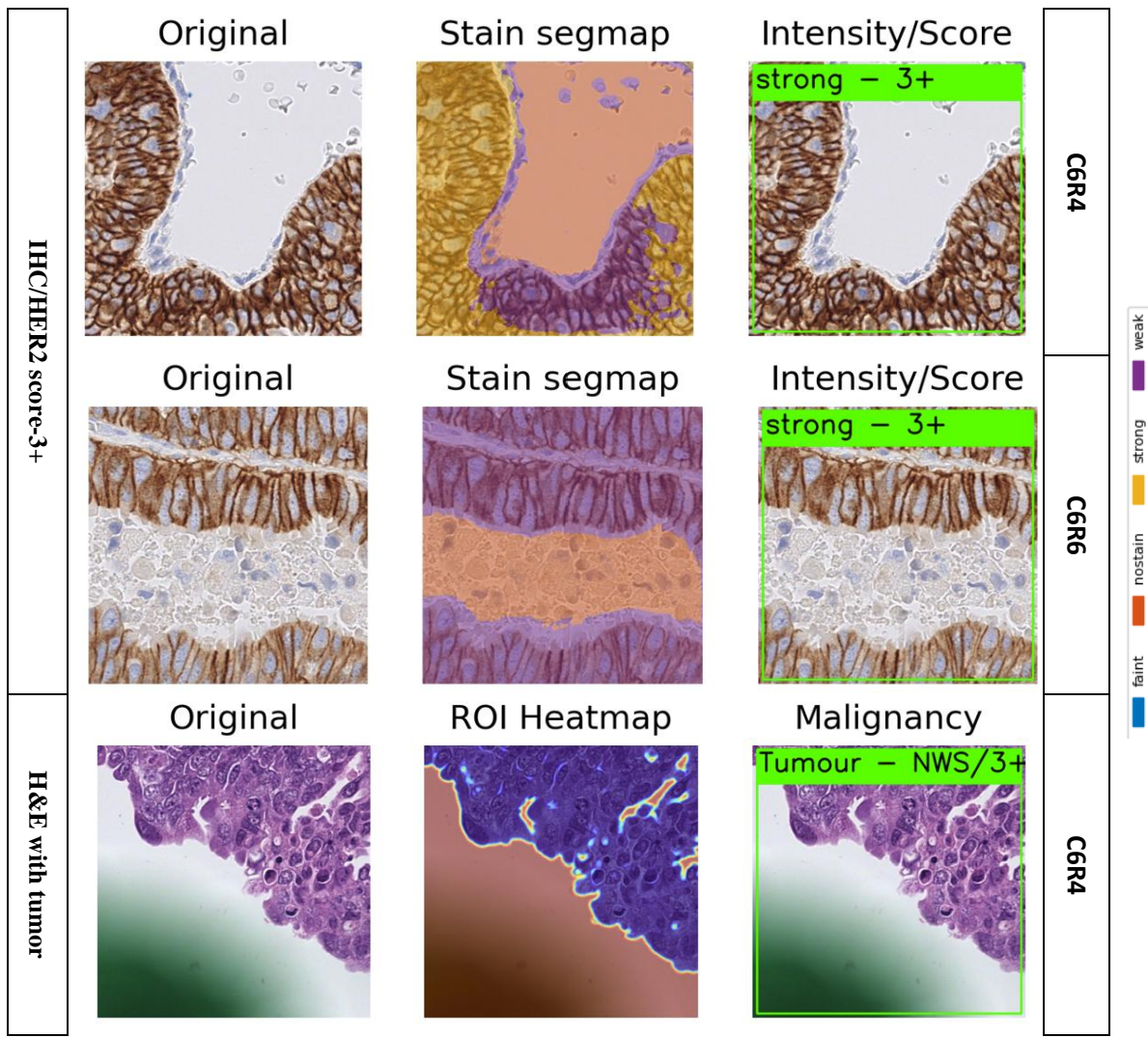

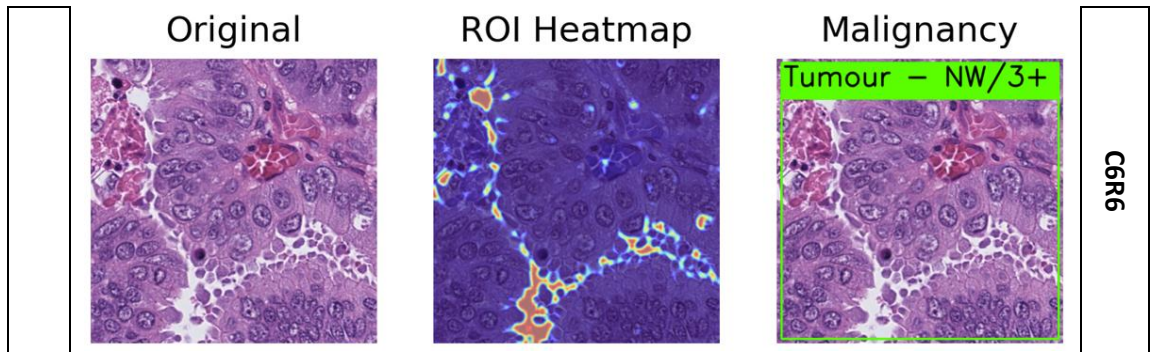

**Figure 12:** The patches of IHC with HER2-score 3+ (strong stain) and H&E with malignant (tumor and artifact), extracted from WSI-case-6, showing segmentation annotation and attention heatmaps

To view the outcome of the annotation of malignancy and stain intensity, patches were collected together into their corresponding ROIs to form a whole block of the WSI from which they were extracted. In Figure 13 (a), a pair of the original IHC ROI and its corresponding heatmap ROI is seen; and in Figure (b) is the original H&E ROI and the corresponding heatmap ROI. Several ROIs are then further combined to form the WSIs for IHC and H&E. This stacking completes the task of WSIs-level processing as described by the proposed pipeline. The implication of this is that the proposed method approaches the targeted task of HER2 scoring from the granular level of image patches. The scoring for the ROIs seen in the figure is 3+ for the IHC, while the H&E is generally determined as having a tumor in the locations of the ROIs.

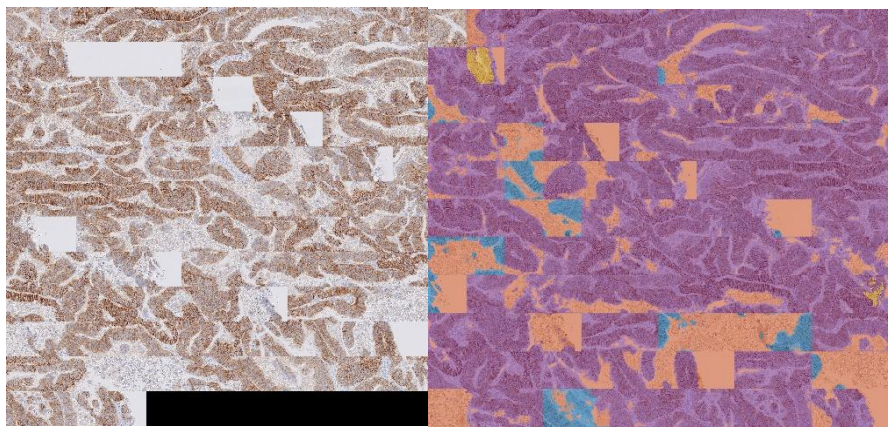

(a)

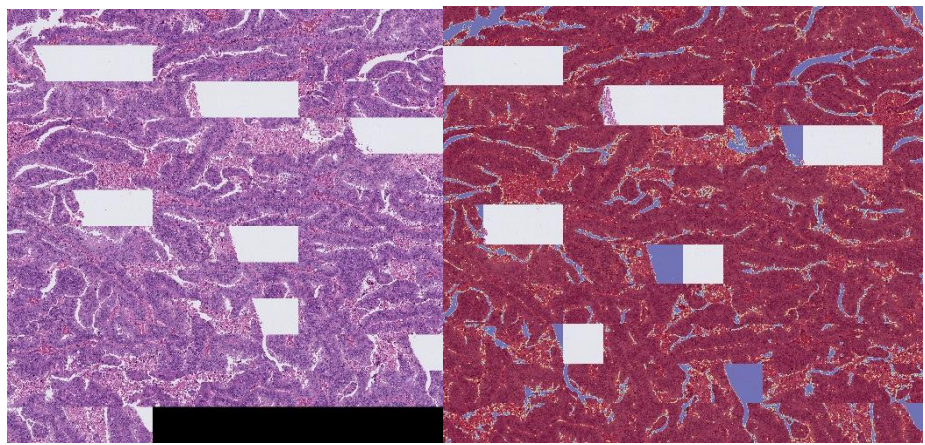

(b)

**Figure 13**: Region of interests (ROIs) of C6r6 extraction from WSIs showing mappings of the attention heatmap of (a) IHC predicted as having a HER2-score of 3+ and (b) H&E by the proposed pipeline

Since the aim of the proposed pipeline is effective HER2 scoring and correctly identifying the intensity of stains in the WSIs, we have also examined these two closely. In Figure 14, we show image patches from IHC that were accurately stained-classified leading to the HER2 scoring. The figure shows that the stain detection is consistent and supportive of the process of scoring the measure of HER2 expression. The implication of this outcome is a confirmation of the applicability of the proposed pipeline to the task of scoring HER2 expression in IHC stain WSIs for the four-way scoring. The four-way scoring described in this study covers the HER2-score of 0 (no stain), HER2-score of +1 (faint stain), HER2-score of +2 (weak stain), and HER2-score of 3+ (strong stain).

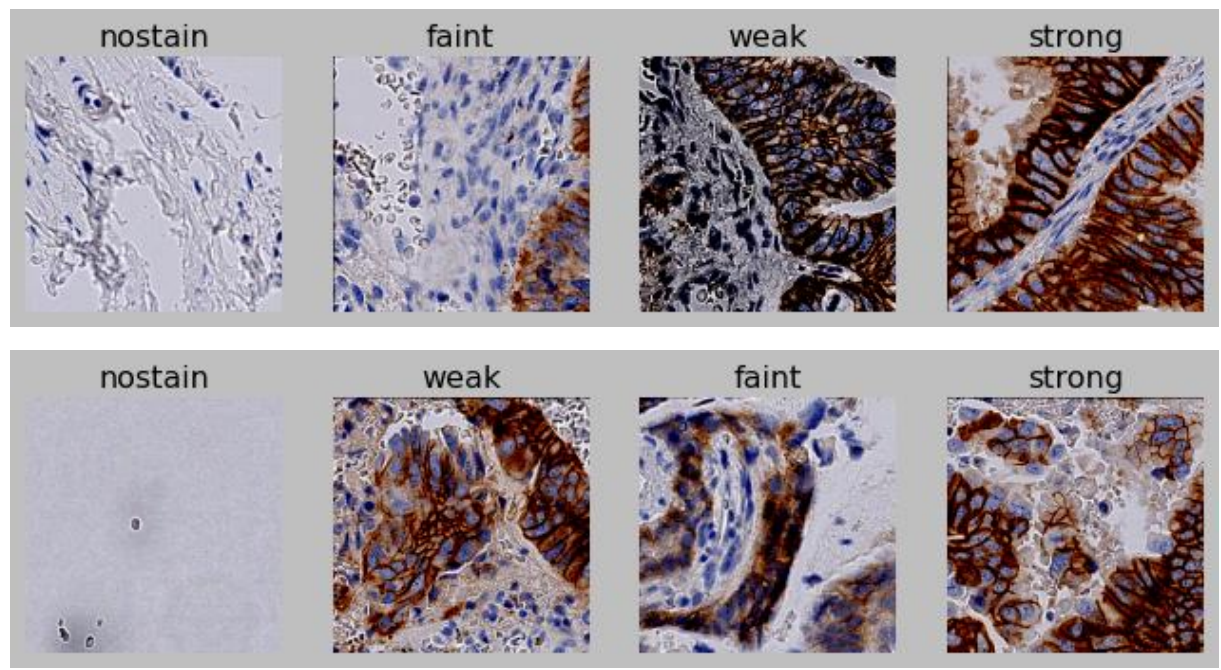

**Figure 14**: Two stripes of some sample patches of IHC images predicted as having HER2-score of 0 (no stain), HER2-score of +1 (faint stain), HER2-score of +2 (weak stain), and HER2-score of 3+ (strong stain)

As a result of the region-based approach to WSI IHC HER2 scoring adopted in this study, we decided to investigate the correctness of the result by comparing the ground-truth with the result our pipeline returned. Using the patches generated from the ROIs, a patch-wise comparison of ground-truth with the predicted is outlined in Table 3. Note that labeling represented by R-tumor denotes a real tumor as seen in the ground-truth, while those with P-tumor represent a predicted tumor. Similarly, all labeling is represented with R-No-stain, R-Fain-stain, R-weak-stain, and R-strong-stain are those presented in the ground-truth listing, while those denoted by P-No-stain, P-Fain-stain, P-weak-stain, and P-strong-stain are those predicted by the proposed pipeline. The result shows that based on the H&E tumor detection, the proposed pipeline competitively demonstrates good performance with the ground-truth on the C6-R1, C6-R6, C6-R2, and C6-R7, while it had some slight drawbacks in C6-R4 and C6-R11. For instance, the C6-R4 ground-truth has 66 patches with tumors and 55 without malignancy. However, our predicted pipeline showed that there are 55 patches with tumors and 67 are normal without tumors. Also, in the case of C6-R11, the ground-truth shows that only one patch has malignancy, while the remaining 143 are normal. However, the proposed pipeline misses the one patch and wrongly classifies it as normal, though it correctly classifies the 143 patches as normal

**Table 3**: A listing showing a comparative outline of the ground-truth against those predicted by the proposed pipeline as applied to patches IHC and H&E from C6R4 and C6R6.

| | | C6 = case 6, C3 = case 3, R# = ROI number | | | | | |
|---|---|---|---|---|---|---|---|
| Task | Image status | C6-R4 | C6-R1 | C6-R6 | C6-R7 | C6-R11 | C6-R2 |
| | R-tumor | 66 | 0 | 131 | 122 | 1 | 0 |
| | P-tumor | 53 | 0 | 131 | 122 | 0 | 0 |
| | R-normal | 55 | 144 | 13 | 22 | 143 | 144 |
| **H&E classification** | P-normal | 67 | 144 | 13 | 22 | 144 | 144 |
| **HER2 Classification** | R-No-stain | 33 | 140 | 7 | 21 | 112 | 140 |
| | R-Fain-stain | 27 | 0 | 26 | 18 | 0 | 0 |
| | R-weak-stain | 28 | 0 | 68 | 81 | 0 | 0 |
| | R-strong-stain | 7 | 0 | 35 | 10 | 0 | 0 |
| | P-No-stain | 14 | 140 | 3 | 11 | 112 | 140 |
| | P-Fain-stain | 18 | 0 | 26 | 18 | 0 | 0 |
| | P-weak-stain | 17 | 0 | 67 | 81 | 0 | 0 |
| | P-strong-stain | 4 | 0 | 35 | 10 | 0 | 0 |

The IHC patches were also compared with the ground-truth, as listed in the table. For instance, the C6-R1, C6-R11, and C6-R2 indicate that the patches were correctly stained and classified, and the HER2 score was also accurately computed. However, the C6-R4, C6-R6, and C6-R7 do show some slight inconsistency in the matching of values with the ground-truth. In summary, the performance of the proposed pipeline in HER2 scoring is very impressive and competes with what the human pathologists have registered in the ground-truth. This subsection has therefore shown the relevance of the study in scoring the HER2 expression in IHC-stained images. In addition, we have shown, using several heatmap H&E patches and ROIs, that the study is capable of classifying abnormal WSIs by detecting regions with malignancy.

### 5.3 Discussion of Findings

The proposed HER2 scoring pipeline reveals some significant outcomes that were observed during the experimentation phase. For instance, the experimental outcome showed that the ViT-based model demonstrated a very good performance in feature learning and extraction. This is important because every image analysis problem strongly depends on an effective feature learning phase to yield an outstanding performance. Hence, the effort was taken to first evaluate the performance of the ViT-based models applied to H&E images and IHC images for tumor detection and stain-intensity classification. Interestingly, the performance is appealing and has contributed to the overall outcome of the proposed pipeline. Therefore, findings from the study confirm the relevance of applying a suitable feature detection mechanism to H&E images and IHC images. Secondly, the result from the study establishes the fact that there is a strong linkage between the stain intensity and with HER2 score and expression. We found out that there is a strong correlation between the stain intensity observed on IHC WSIs and patches with the quantification of HER2 expression and scoring for the same IHC WSI/patches. This therefore confirms that our approach to tie the stain-intensity classification to the HER2 scoring is relevant. Moreover, the four-way (0, 1+, 2 + and 3 +) scoring model applied to this study consistently reported a strong correlation with the four different patterns of stains identified in the study. Therefore, findings from the approach demonstrate that both pathologists and AI-driven models can continue to leverage the pattern of stain in IHC WSIs to quantify the HER2 expression and to achieve the scoring.

The conventional conceptual knowledge of the decomposed problem-solving paradigm often requires that a big task be decomposed into smaller tasks. The aim is to solve the smaller tasks and then assemble the solution into the bigger solution. While it may be good to process WSI, our approach establishes the fact that patch-wise processing and analysis of WSIs are promising are can be accurate. In this study, the search for tumor and stain-intensity classification was achieved first at the granular levels of WSIs patches. We found that it is easier to accurately analyse the pixels at such granularity with reduced computational resources compared with analyzing the WSI. As a result, findings from this study confirm the importance of a patch-wise approach to WSIs analysis, both for H&E and IHC. Moreover, the application of the segmentation model to effectively annotate and segment different classes of stains on a patch, ROIs, or WSIs. In this study, the application of the staining segmentation was strictly to the IHC patches. However, we discovered that such annotation and segmentation will also be very useful for the localization of tumors and other abnormalities in H&E WSIs/patches. As a result, such localization will support the explainability of the outcome from such AI-driven pipeline proposed in this study. This therefore provides a complete pipeline that classifies, scores HER2 expression, and accurately annotates with a segment map the location of the malignancy and stains in H&E and IHC WSIs, respectively. In summary, the main aim of this study is to score the HER2 expression in IHC images and WSI and to jointly classify the stain intensity in these images. In addition to these core tasks, the study also aims at detecting tumors in patches of WSIs and at the

WSI level. Outcome from the experimentation and results presentation stages demonstrates acceptable performance.

Although our proposed pipeline was tested on a privately curated dataset of H&E and IHC, we thought that it was also important to comparatively show how it performs besides some state-of-the-art (SOTA) methods in HER2 expression scoring. In Table 4, an outline of some similar methods is listed and showing their methods and performance. Interestingly, we found the works of [28], [27], and [46] are significantly similar to our proposed method based on the use of IHC and H&E inputs. While these studies returned the performance of accuracy of 0.87, AUC of 0.783, and AUC of 90%, our proposed model outperformed with Accuracy, precision, F1-score, sensitivity, and recall of 1.0 on tumor; precision, F1-score, sensitivity, recall, and accuracy of 0.982, 0.991, 1.0, 1.0, and 0.9400, respectively on four score classification (0, 1+, 2 + and 3 +). We found the approach of contrast learning [40], [27], and correlation attention learning [31], to be interesting. However, our proposed method yet demonstrates some measure of superiority in terms of performance. The implication is that the method proposed I this study is more effective in HER2 scoring with high classification accuracy on H&E and IHC staining detection. Moreover, we found most of the other SOTA using the ViT-based approach to achieve feature learning and extraction.

Table 4: Comparative performance analysis with SOTA methods for HER2 status scoring

| Ref | Method | Performance |
| --- | --- | --- |
| [28] | HE-HER2Net: A transfer learning approach with Grad-CAM for explainability | accuracy (0.87), precision (0.88), recall (0.86), and AUC score (0.98) |
| [40] | A weak-supervised model with MoCo-v2 contrastive learning | Area Under the Curve (AUC) of 0.85 |
| [27] | WSMCL: weakly supervised multi-modal contrastive learning on H&E and IHC | ACC, Macro-AUC, Weighted-AUC, Macro-F1, Weighted-F1 of 0.783, 0.941, 0.945, 0.762, and 0.781, respectively |
| [45] | A 3-stage model for weakly localizing features, an attention module, and HER2 expression level proximity computation | AUC of 0.9202, precision of 0.922, sensitivity of 0.876, and specificity of 0.959 |
| [47] | Workflow using Vit on a hybrid of Next-Generation Sequencing (NGS) data and IHC staining images | Accuracy is 93.1%, and AUC is 0.841 |
| [42] | end-to-end ConvNeXt network utilizing low-resolution IHC images | AUC, F1, and accuracy of 91.79%, 83.52%, and 83.56% respectively |
| [46] | ViT-based pipeline for HER2 scoring using H&E | AUC is 90% |
| [43] | DenseNet201, GoogleNet, MobileNet_v2, and ViT models were applied for classification, and a random forest for HER2 score prediction | Accuracy of 92.6% and 91.15% for malignancy and scoring, on patch score classification. |
| [31] | Feature Learning framework based on Correlational Attention Neural Network (Corr-A-Net) for HER2-scoring | Accuracy of 0.93 and AUC of 0.98, mean effective confidence (MEC) score of 0.85 |
| [32] | Vision Transformer (ViT) model based on dynamic contrast-enhanced MRI (DCE-MRI) | AUCs of 0.86, 0.80, and 0.79 in the classification of HER2-low and HER2-positive |
| **This study** | ViT-based tumor and stain classifier, and stain segmentation | Accuracy, precision, F1-score, sensitivity, and recall of 1.0 on tumor; precision, F1-score, sensitivity, recall, and accuracy of 0.982, 0.991, 1.0, 1.0, and 0.9400, respectively on four score classification (0, 1+, 2 + and 3 +). |

6.  Conclusion

In this study, an end-to-end pipeline method is proposed based on multilayered ViT-based models. The study focuses on the classification of WSIs representative of H&E and IHC. We proposed a creative way of achieving tumor classification on a patch-wise level and ROIs-based level in H&E WSI. Similarly, the IHC WSIs were processed to extract patches for further analysis in detecting the intensity of stains. Both the classification models for malignant detection and stain intensity were supportive in scoring the measure of HER2 expression in IHC WSIs. We applied a novel approach to correspond the tumor-defected regions of H&E tissue with an exact region in the matching IHC WSIs. Moreover, he proposed a study that creatively adapted a segmentation transformer to accurately achieve pixel-level categorization and segmentation of different classes of stains, suggesting HER2 status. The segmentation model aims to help accurately map regions with varied stain intensities, with the hope of supporting the HER2 scoring process. Two separate ViT-based models were proposed for the classification task on tumor identification and HER2 stain-intensity-status computation. We privately curated datasets to experiment with the proposed model. After exhaustive experimentation, the complete pipeline was comparatively analyzed for applicability to the core task of HER2 scoring. In addition, the performance of the proposed pipeline was compared with that of several state-of-the-art (SOTA) methods. The Result obtained showed that the four-way HER2 scoring (0, 1+, 2 + and 3 +) was at 0.940 and 1.0 for stain-intensity and malignant classification accuracy. In the future, we plan to investigate the suitability of a hybrid ViT-based model capable of addressing the peculiarity of the images. We also recommend that the localization of malignant regions have a similar segmentation mapping as seen in our stain-intensity mapping. This will provide a clearer visualization of the segmented mapping annotated on H&E and IHC WSIs. I addition, a robust way of processing WSIs without the need for patch-wise extraction will prove helpful, though this will require more computational resources and better neural architecture.